\documentclass[11pt, a4paper, logo, twocolumn, copyright]{googledeepmind}

\pdftrailerid{redacted}

\makeatletter
\renewcommand\bibentry[1]{\nocite{#1}{\frenchspacing\@nameuse{BR@r@#1\@extra@b@citeb}}}
\makeatother

\usepackage{kantlipsum, lipsum}
\usepackage{dsfont}
\usepackage{dm-colors}

\usepackage{algorithm}
\usepackage{algpseudocode}
\usepackage{adjustbox}
\usepackage{multirow}
\usepackage{nicefrac}
\usepackage{tcolorbox}
\usepackage{xcolor} % Include the color package

\newcolumntype{R}[2]{%
    >{\adjustbox{angle=#1,lap=\width-(#2)}\bgroup}%
    l%
    <{\egroup}%
}

\usepackage[authoryear, sort&compress, round]{natbib} 
\usepackage{comment}

\newcommand{\revise}[1]{{\textcolor{magenta}{#1}}}
\newcommand{\myparagraph}[1]{\vspace{-5pt}\paragraph{#1}}

\definecolor{myblue}{HTML}{0153d6} % Replace 0000FF with the hex code for your blue
\definecolor{myred}{HTML}{ff409b} % Replace FF0000 with the hex code for your red

\newcommand{\myobs}[1]{
  \vspace{5pt}\noindent\textbf{\textcolor{myblue}{#1}}
}

\usepackage{pifont}% http://ctan.org/pkg/pifont

\newcommand{\myfn}[1]{\textcolor{purple}{#1}}
\newcommand{\lsgd}{Local-SGD}

\definecolor{commentcolor}{RGB}{110, 154, 155}

\algblock{ParallelFor}{EndParallelFor}
\algnewcommand\algorithmicparallelfor{\textbf{parallel for}}
\algnewcommand\algorithmicendparallelfor{\textbf{end parallel for}}
\algrenewtext{ParallelFor}[1]{\algorithmicparallelfor\ #1\ \algorithmicdo}
\algrenewtext{EndParallelFor}{\algorithmicendparallelfor}

\usepackage{subcaption}
\usepackage{url}

\graphicspath{{figures/}}

\title{Asynchronous Local-SGD Training for Language Modeling}

\correspondingauthor{bliu@cs.utexas.edu}

\reportnumber{} % Leave blank if n/a 

\renewcommand{\today}{} 

\author[*1]{Bo Liu}
\author[2]{Rachita Chhaparia}
\author[2]{Arthur Douillard}
\author[3]{Satyen Kale}
\author[2]{Andrei A. Rusu}
\author[2]{Jiajun Shen}
\author[2]{Arthur Szlam}
\author[2]{Marc'Aurelio Ranzato}
\affil[*]{Work done as an intern at Google DeepMind}
\affil[1]{The University of Texas at Austin}
\affil[2]{Google DeepMind}
\affil[3]{Google Research}

\begin{abstract}
Local stochastic gradient descent (\lsgd{}), also referred to as federated averaging, is an approach to distributed optimization where each device performs more than one SGD update per communication.  
This work presents an empirical study of {\it asynchronous} \lsgd{} for training language models; that is, each worker updates the global parameters as soon as it has finished its SGD steps.  We conduct a comprehensive investigation by examining how worker hardware heterogeneity, model size, number of workers, and optimizer could impact the learning performance.  We find that with naive implementations, asynchronous \lsgd{} takes more iterations to converge than its synchronous counterpart despite updating the (global) model parameters more frequently. We identify momentum acceleration on the global parameters when worker gradients are stale as a key challenge.  We propose a novel method that utilizes a delayed Nesterov momentum update and adjusts the workers' local training steps based on their computation speed. This approach, evaluated with models up to 150M parameters on the C4 dataset, matches the performance of synchronous \lsgd{} in terms of perplexity per update step,  and significantly surpasses it in terms of wall clock time.
\end{abstract}

\begin{document}

\maketitle

\section{Introduction} \label{sec:intro}
Large language models (LLMs) have revolutionized many applications, transforming the way machines interact with human language. The cornerstone of this revolution is training these models at massive scale. To manage such large-scale training in reasonable amounts of time, it has been necessary to distribute computations across multiple devices.  However, the standard approaches to this distributed training uses co-located devices with fast interconnects.  

One might hope to be able to
%As the field of AI progresses, it is anticipated that 
effectively harness a broader range of computational resources, perhaps geographically distant from each other, in order to build even more powerful large models. However, utilizing numerous distant devices faces a significant hurdle: communication latency. When devices focus solely on computing gradients before sending them back to a central server, the communication time can overshadow the computation time, creating a bottleneck in efficiency.
\begin{figure}[h!]
    \centering
    \includegraphics[width=1\columnwidth]{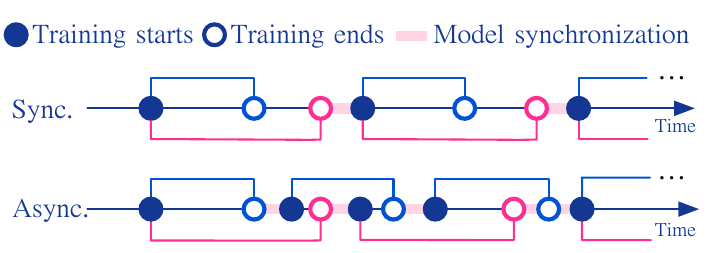}
    \caption{Illustration of async. v.s. sync. training with 2 workers (in \textcolor{myblue}{blue} and \textcolor{myred}{red}). Sync. training suffers from the straggler effect, while async. training reduces the idling time of the fast worker.}
    \label{fig:example_sync_vs_async}
\end{figure}
\begin{figure*}[t!]
    \centering
    \includegraphics[width=0.95\textwidth]{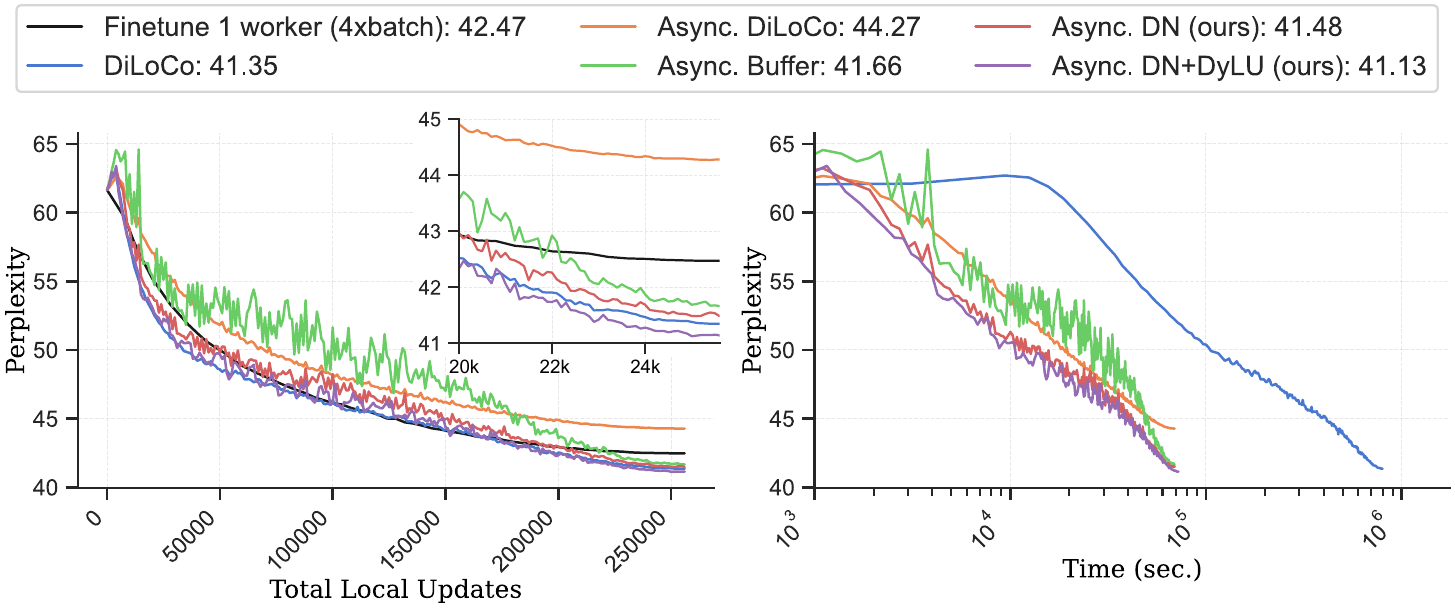}
    \caption{Comparative evaluation of language models using sync. and async. \lsgd{} methods with 4 heterogeneous workers on a 20M parameter model. The state-of-the-art sync. \lsgd{} method, DiLoCo~\citep{douillard2023diloco}, employs \texttt{AdamW} and \texttt{Nesterov} momentum as the worker-side and server-side optimizers, respectively. This optimizer combination remains the strongest for async. \lsgd{} training (See Figure~\ref{fig:all_optimizer}), yet underperforms DiLoCo significantly. By integrating Delayed Nesterov (DN) (Algorithm~\ref{alg:delayed_nesterov}) for outer optimization and Dynamic Local Updates (DyLU) (Section~\ref{sec:dylu}), we significantly bridge the performance gap in terms of perplexity versus updates between sync. and async. training in language modeling. Moreover, the proposed method significantly surpasses DiLoCo in terms of perplexity versus wall clock time.}
    \label{fig:async_delayed_nesterov_very}
\end{figure*}

Local Stochastic Gradient Descent (\lsgd{}) is a collection of optimization methods that can reduce communication bottlenecks.\footnote{The term \lsgd{}, sometimes also known as Federated Average (FedAvg), is used here to emphasize its roots in distributed optimization, where users have control over data allocation to different workers.} These methods involve each device performing multiple local gradient steps before syncing their parameter updates with a parameter server. While \lsgd{} enhances training efficiency by reducing communication frequency, it can suffer from the \emph{straggler effect} caused by heterogeneous devices. For instance, faster devices are idle waiting for slower ones to catch up, undermining the overall efficiency of the system. Moreover, all devices are forced to communicate at the same time requiring high bandwidth connection with the parameter server.  Asynchronous \lsgd{} presents a more viable solution (illustrated in \autoref{fig:example_sync_vs_async}), as it allows the server to update the model as soon as the updates of a worker are available, thereby enhancing computational utilization and minimizing communication bandwidth requirements.

In this study, we explore the viability of asynchronously training LLMs using \lsgd{}.  We expand upon previous works that have attempted to alternate steps on subsets of workers or randomly drop certain subset of workers during synchronous \lsgd{}~\citep{ryabinin2021moshpit, douillard2023diloco}. 
%However, to our knowledge, async. \lsgd{} has not been thoroughly investigated in the context of training Large Language Models (LLMs). 
The main content is structured into three parts:
\myparagraph{1. Framework (Section~\ref{sec:async_framework}).} The first part introduces our high-level design for the asynchronous training framework. We discuss how each worker determines which data shard to train on, for how many steps, with what learning rates, and how the server updates models asynchronously.
\myparagraph{2. Optimization Challenge (Section~\ref{sec:challenge}).} In the second part, we conduct an empirical study of various existing optimization strategies suitable for asynchronous \lsgd{}. This includes both worker-side optimization (inner optimization) and server-side optimization (outer optimization). We uncover a key challenge in utilizing momentum effectively. Notably, while adaptive momentum methods generally accelerate convergence of both inner and outer optimizations, their efficacy in asynchronous \lsgd{} is comparatively reduced when both optimizations employ momentum techniques, especially when contrasted with the synchronous implementation.
\myparagraph{3. Proposed Solutions (Section~\ref{sec:method}).} 
We introduce two simple and effective techniques: the Delayed Nesterov momentum update (DN) and Dynamic Local Updates (DyLU). These techniques, when combined and evaluated on training language model, allow asynchronous \lsgd{} to approach synchronous \lsgd{} in terms of perplexity versus the total number of local updates, and further improve asynchronous \lsgd{} vs. synchronous \lsgd{} in terms of perplexity versus wall-clock, as detailed in Figure~\ref{fig:async_delayed_nesterov_very}.

\revise{
%\myparagraph{Summary of Contribution:} \textbf{1)} We identified an optimization challenge when applying momentum to the outer optimizer during asynchronous \lsgd{} training for language models. \textbf{2)} To address this, we introduce two simple and effective techniques: the Delayed Nesterov momentum update (DN) and Dynamic Local Updates (DyLU). These techniques, when combined, significantly reduce the performance gap between asynchronous and synchronous \lsgd{} in language model training. This improvement is evident in terms of perplexity versus the total number of local updates, as detailed in Figure~\ref{fig:async_delayed_nesterov_very}.
}

\section{Background}
In this study, we focus on the distributed optimization of shared model parameters $\theta$ across $k$ data shards, denoted as $\mathcal{D} = \{ \mathcal{D}_1, \dots, \mathcal{D}_k\}$, with $k$ workers.\footnote{We assume the number of workers ($k$) equals the number of data shards, though our methods are also applicable when there are fewer workers than data shards.} The primary goal is described by the following equation:
\begin{equation}
\min_\theta \sum_{i=1}^k \frac{|\mathcal{D}_i|}{\sum_j |\mathcal{D}_j|} \mathbb{E}_{x \sim \mathcal{D}_i} \big[ \ell(x; \theta) \big],
\end{equation}
where $\ell(\cdot; \theta)$ represents the loss function (for instance, cross entropy loss for next token prediction in language modeling), and $|\cdot|$ indicates the set size.
\begin{algorithm}[t!]
\caption{DiLoCo Algorithm (synchronous)} \label{alg:local_sgd}
\begin{algorithmic}[1]
\Require Initial pretrained model $\theta^{(0)}$
\Require $k$ workers
\Require Data shards $\{\mathcal{D}_1, \dots, \mathcal{D}_k\}$
\Require Optimizers $\texttt{InnerOpt}$ and $\texttt{OuterOpt}$

\For{\texttt{outer step $t = 1 \ldots T$}}
  
  \ParallelFor{\texttt{worker $i = 1 \ldots k$}}
    \State $\theta_i^{(t)} \gets \theta^{(t-1)}$
    \For{\texttt{inner step $h = 1 \ldots H$ }}
        \State $x \sim \mathcal{D}_i$
        \State $\mathcal{L} \gets f(x, \theta_i^{(t)})$
        \State $\theta_i^{(t)} \gets \texttt{InnerOpt}(\theta_i^{(t)}, \nabla_\mathcal{L})$
    \EndFor
    \State $\delta_i^{(t)} = \theta^{(t-1)} - \theta_i^{(t)}$
  \EndParallelFor
  \State $\Delta^{(t)} \gets \frac{1}{k} \sum_{i=1}^k \delta_i^{(t)}$\Comment{\textcolor{gray}{outer gradient}}
  \State $\theta^{(t)} \gets \texttt{OuterOpt}(\theta^{(t-1)}, \Delta^{(t)})$
\EndFor
\end{algorithmic}
\end{algorithm}

We extend the definition of \lsgd{} in this work to include not just the original \lsgd{} method, but also its variants that incorporate advanced optimization techniques. We particularly focus on DiLoCo~\citep{douillard2023diloco}, which sets the standard for synchronous \lsgd{} in language modeling. DiLoCo's methodology is detailed in Algorithm~\ref{alg:local_sgd}. Each worker $i$ performs $H$ local updates using an \emph{inner optimizer} on their data shard $\mathcal{D}_i$ before sending the parameter change (pseudo-gradient) $\delta^{(t)}_i = \theta^{(t-1)} - \theta^{(t)}_i$ back to the server. The server then computes the aggregated outer gradient $\Delta^{(t)} = \frac{1}{k} \sum_{i=1}^k \delta^{(t)}_i$, and applies an \emph{outer optimizer} with $\Delta^{(t)}$ to update $\theta$. A key insight from DiLoCo is the optimal use of \texttt{AdamW} and \texttt{Nesterov Momentum} as the best inner and outer optimizers, respectively.
\section{Async. Local-SGD Framework} \label{sec:async_framework}
This section outlines the asynchronous \lsgd{} pipeline design, where we assume a central server controls all workers and asynchronously aggregates their updates.

\myparagraph{Data Shard Sampling} Unlike in the federated learning setting where each device is attached to its own data, in distributed optimization, the user has the right to choose which data shard is assigned to which worker, even dynamically. To balance the learning progress on different data shards (as workers are heterogeneous), whenever a worker is ready to start a new local optimization round, we sample a data shard inversely proportional to its ``learning progress". Specifically, define $n_i$ as the number of learned data points in $\mathcal{D}_i$, then we sample a shard $i_\text{sampled}$ according to:
\begin{equation}
\begin{split}
i_\text{sampled} &\sim p, \\
\text{where}~p_i &\propto \text{max}(\frac{|\mathcal{D}_i|}{\sum_j |\mathcal{D}_j|} - \frac{n_i}{\sum_j n_j}, 0). 
\end{split}
\label{eq:data_sample}
\end{equation}
In other words, we sample a data shard only when it is ``under-sampled" (i.e., $\frac{n_i}{ \sum_j n_j} \leq \frac{|\mathcal{D}_i|}{\sum_j |\mathcal{D}_j|}$). The degree to which a shard is under-sampled determines its sampling rate. By doing so, we ensure that the data shard with slower progress is more likely to be sampled for training, therefore balancing the learning progress across shards.

\myparagraph{Learning Rate Scheduling} In contrast to synchronous training methods like DiLoCo, asynchronous training can lead to uneven progress across different data shards, especially when workers are allowed varying numbers of training steps. This  raises the question of how to effectively schedule learning rates. In our approach we assign each data shard its own learning rate schedule. Specifically, we implement a linear warmup combined with a cosine learning rate decay, where $T$ represents the target total training iterations for each data shard:
\begin{equation}
\eta_t = \begin{cases}
 t \eta_\text{max}/t_\text{warmup} & t < t_\text{warmup} \\
 \eta_\text{min} + 0.5 (\eta_\text{max} - \eta_\text{min}) \\ ~~~~\big( 1 + \cos\big(\frac{t - t_\text{warmup}}{T - t_\text{warmup}}\pi \big)\big) & t \geq t_\text{warmup}.
\end{cases}
\label{eq:lr}
\end{equation}
In practice, asynchronous training may conclude with different final iteration counts ($t_\text{end}$) for each data shard. Since we cannot predetermine $t_\text{end}$ due to the unpredictability of asynchrony, we set the minimum learning rate ($\eta_\text{min}$) to a small positive value. This ensures continued progress even if $t$ exceeds $T$. Additionally, we adjust $T - t_\text{warmup}$ to be non-negative and ensure that the ratio $\frac{t - t_\text{warmup}}{T - t_\text{warmup}}$ remains within the range of $[0, 1]$. This helps maintain effective learning rate adjustments throughout the training process.

\begin{figure}[h!]
    \centering
    \includegraphics[width=0.8\columnwidth]{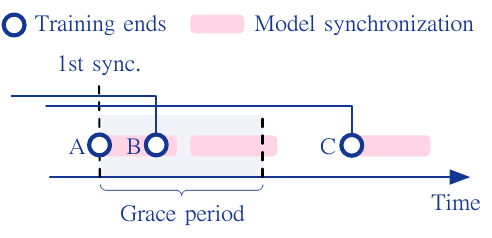}
    \caption{We consecutively synchronize the update from B after we synchronize A because B finishes its training after A but before the end of the grace period. A and B will therefore use the same server model to start the new training jobs, while C will start its own grace period. }
    \label{fig:grace_period}
\end{figure}

\myparagraph{Grace Period for Model Synchronization} In asynchronous training, the completion time of each worker's tasks can vary. For example, if worker B completes training shortly after worker A, it might be beneficial for A to wait briefly until the server processes updates from both workers before receiving the updated model for its next training task. However, this waiting period should be minimal and occur only when necessary. Specifically, if no other worker completes its task within the grace period while worker A is synchronizing with the server's model, A should promptly commence its new training task using the server's current model.
For a visual representation of this process, please refer to Figure~\ref{fig:grace_period}.

\myparagraph{Asynchronous Task Scheduling} In Algorithm~\ref{alg:async_pipeline}, we present the asynchronous task scheduling pipeline. Throughout the algorithm, we use $\tau$ to denote the actual wall clock time and $t$ to denote model updates. In \textbf{line 1-4}, we initialize the model, total local updates $t_\text{local}$, and the list of workers $\mathcal{W}$ and the completed workers $\mathcal{W}_\text{completed}$. In \textbf{line 5}, we start the first training job for all workers with the initial model parameter $\theta^{(0)}$. Note that the \myfn{train}() function implements the data sampling technique and performs the learning rate scheduling mentioned before. In \textbf{line 6}, we reset the starting time of the grace period $\tau_\text{sync}$ to $\infty$. This is because we want to synchronize with a worker only when its completion time is within $\tau_\text{sync} + \tau_\text{grace}$. The main asynchronous \lsgd{} training loop is provided in \textbf{line 6-19}. Within the loop, we first attempt to get a completed worker $w$ (\textbf{line 7}). We retrieve the earliest completed worker that we have not yet processed yet, as long as its completion time is still within the grace period (e.g., $w$.completed\_time $\leq \tau_\text{sync} + \tau_\text{grace}$). If no such workers exist, \myfn{get\_worker}() will return null. In \textbf{line 10-15} where such a worker $w$ is found, we synchronize its update with the server model $\theta$. In \textbf{line 17-20} when no such workers are found, we assign new training jobs for all completed workers and empty the list of completed workers. For the detailed pseudocode of the \myfn{train}() and \myfn{get\_worker}() functions, please refer to Appendix~\ref{sec:pseudo}.
In practice, for the sake of reproducibility of research, we implement a \emph{determininistic} version of Algorithm~\ref{alg:async_pipeline} with faked training time based on real-world device statistics. We validate the correctness of the training pipeline by simulating synchronous updates using the asynchronous framework.

\begin{algorithm}[t!]
\caption{Async. Local-SGD Task Scheduling.} \label{alg:async_pipeline}
\begin{algorithmic}[1]
\Require Initial pretrained model $\theta^{(0)}$
\Require $k$ workers
\Require Grace period $\tau_\text{grace}$
\Require Total local updates $t_\text{max}$
\State \texttt{$t_\text{local} = 0$}
\State \texttt{$\theta \leftarrow \theta^{(0)}$}
\State \texttt{$\mathcal{W}$ = [\myfn{init\_worker}() for $i$ in [$k$]]}
\State \texttt{$\mathcal{W}_\text{completed}$ = []}
\State \texttt{\myfn{train}($\mathcal{W}$, $\theta$)}
\State \texttt{$\tau_\text{sync} = \infty$} \Comment{\textcolor{gray}{start of the grace period}}
\While{\texttt{$t_\text{local}  < t_\text{max}$}}
    \State \texttt{$w$ = \myfn{get\_worker}($\mathcal{W}, \tau_\text{grace}, \tau_\text{sync}$)}
    \State \Comment{\textcolor{gray}{get a completed worker}}
    \If{\texttt{$w$ exists}}
        \State \Comment{\textcolor{gray}{synchronize the update with server}}
        \State $\tau_\text{sync}$ = min($\tau_\text{sync}$, $w$.completed\_time)
        \State \texttt{$\theta \leftarrow$ \myfn{sync}($\theta$, $w$.update)}
        \State \texttt{$\mathcal{W}_\text{completed}$.add($w$)}
        \State \texttt{$t_\text{local}$ += $w$.local\_updates}
    \Else
        \State \Comment{\textcolor{gray}{assign jobs for completed workers}}
        \State $\tau_\text{sync} = \infty$
        \State \texttt{\myfn{train}($\mathcal{W}_\text{completed}$, $\theta$)}
        \State \texttt{$\mathcal{W}_\text{completed}$ = []}
    \EndIf 
\EndWhile
\end{algorithmic}
\end{algorithm}

\section{Optimization Challenge} \label{sec:challenge}
\begin{figure}[h!]
    \centering
    \includegraphics[width=0.45\textwidth]{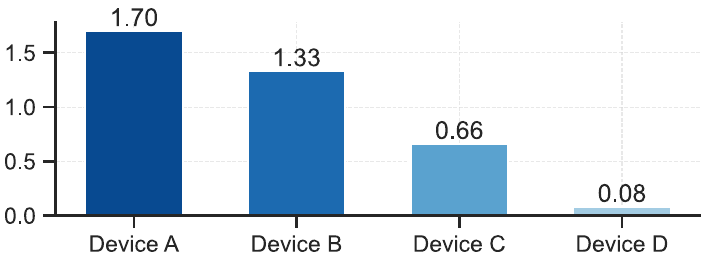}
    \caption{Steps per second for each device.}
    \label{fig:very_staleness_device_speed}
    \vspace{-10pt}
\end{figure}

\begin{figure}[h!]
    \centering
    \includegraphics[width=0.5\textwidth]{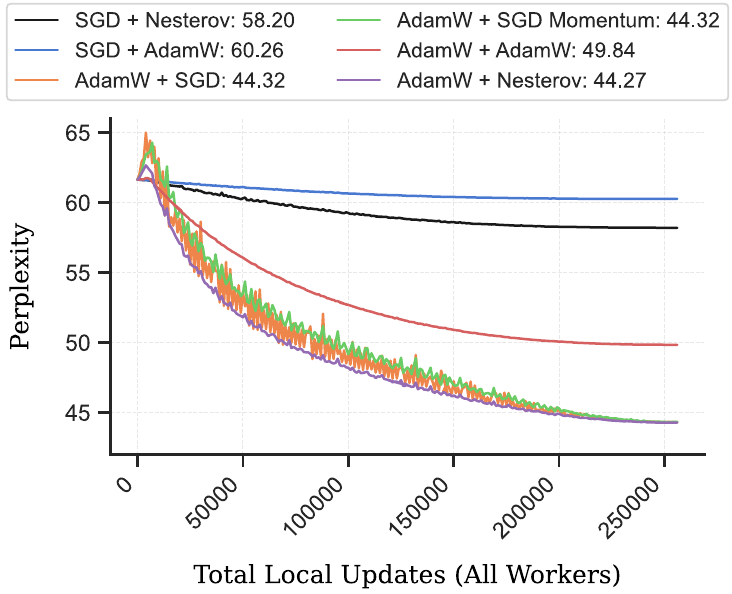}
    \caption{Performance of using different combinations of inner and outer optimizers for asynchronous \lsgd{} training on a 20M language model with 4 workers.}
    \label{fig:all_optimizer}
\end{figure}

\paragraph{Effect of \texttt{InnerOpt} + \texttt{OuterOpt}} To study how optimization affects the language modeling performance in asynchronous \lsgd{}, we first experiment with different combinations of the inner and outer optimizers (we use A+B to denote A and B as the inner and outer optimizer, respectively): \texttt{SGD}+\texttt{Nesterov}, \texttt{SGD}+\texttt{Adam}, \texttt{AdamW}+\texttt{SGD}, \texttt{AdamW}+\texttt{SGD Momentum}, \texttt{AdamW}+\texttt{Adam}, \texttt{AdamW}+\texttt{Nesterov}. The hyperparameters for each combination are tuned separately, for AdamW as \texttt{InnerOpt} we kept the default values. We assume there are $k=4$ workers, whose device speed is shown in Figure~\ref{fig:very_staleness_device_speed}. Then we apply asynchronous \lsgd{} finetuning on a 20M-parameter language model for $64{,}000$ steps per worker ($256{,}000$ local steps in total), where the initial model checkpoint was pretrained for $24{,}000$ steps with \texttt{Adam} without distributed training. We choose finetuning with \lsgd{} as it has been observed that \lsgd{} style methods work well in finetuning but is less efficient from scratch~\citep{lin2018don}, though others have also observed that \lsgd{} works well even for training from scratch~\citep{douillard2023diloco}. The learning rate scheduling and task scheduling follow the procedures described in Section~\ref{sec:async_framework}. We use inner steps = $50$ across all workers in all experiments by default. The result is shown in Figure~\ref{fig:all_optimizer}.

\myobs{Observation} The analysis reveals that combining \texttt{AdamW} as the inner optimizer with \texttt{Nesterov} momentum as the outer optimizer yields the best results, aligning with findings from synchronous training, like the DiLoCo method. Notably, using \texttt{AdamW} as the outer optimizer is less effective. This may be because \texttt{AdamW}, derived from \texttt{Adam}, introduces a normalization effect, which can be counterproductive in \lsgd{} where pseudo-gradients tend to be larger than true gradients, potentially slowing convergence. When \texttt{AdamW} is used in the inner optimization, \texttt{SGD}, \texttt{SGD Momentum}, and \texttt{Nesterov} show comparable performance. However, \texttt{Nesterov} not only stabilizes the learning curve but also slightly improves final performance. This can be attributed to its update mechanism (here we abuse the notation and let $t$ denote $t_\text{server}$):
\begin{equation}
\label{eq:nesterov}
\begin{split}
m_{t+1} &= \beta m_t + g_t \\
\theta_{t+1} &= \theta_t - \epsilon \big(\beta^2 m_t + (1 + \beta) g_t\big),
\end{split}
\end{equation}
where $\epsilon$ is the learning rate, $m_t$ is the momentum, $g_t$ the gradient at time $t$, and $\beta \in (0, 1)$ the decay factor. The key difference between \texttt{Nesterov} and \texttt{SGD Momentum} is in how \texttt{Nesterov} adjusts the weightings, reducing the momentum component ($\beta^2$ instead of $\beta$) and increasing the gradient component ($1 + \beta$ instead of $1$). This suggests that momentum plays a crucial yet intricate role in \lsgd{}.

\begin{figure}[h!]
\centering
\includegraphics[width=0.5\textwidth]{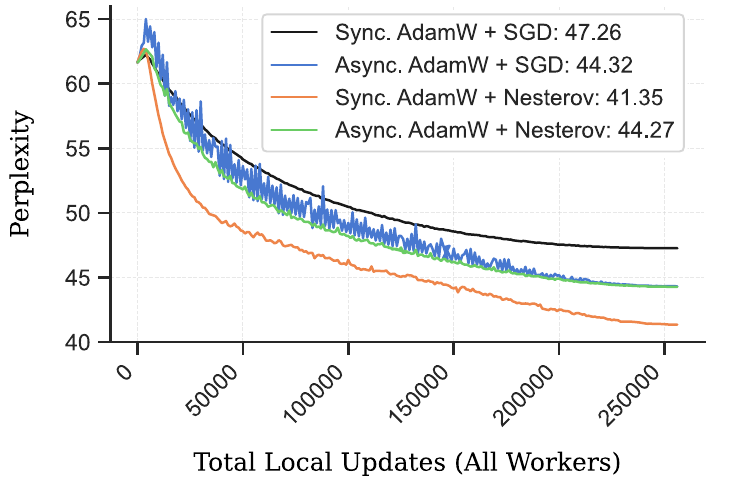}
\caption{Comparison of \texttt{AdamW}+\texttt{SGD} and \texttt{AdamW}+\texttt{Nesterov} in both synchronous and asynchronous \lsgd{} training.}
\label{fig:nesterov_or_not}
\end{figure}

\myparagraph{Momentum in the \texttt{OuterOpt}} To delve deeper into the momentum term's impact on the outer optimizer, we conducted comparative analyses between \texttt{AdamW}+\texttt{SGD} and \texttt{AdamW}+\texttt{Nesterov} under both synchronous and asynchronous training settings. These experiments were carried out under identical conditions as described earlier. The results are reported in Figure~\ref{fig:nesterov_or_not}.

\myobs{Observation} The figure clearly shows that in asynchronous \lsgd{}, \texttt{AdamW}+\texttt{SGD}, which lacks a momentum term, leads to better final perplexity and learning efficiency than its synchronous counterpart. However, incorporating \texttt{Nesterov} momentum into the \texttt{OuterOpt} significantly boosts the performance of synchronous \lsgd{}, outperforming the asynchronous version. It's noteworthy that asynchronous \texttt{AdamW}+\texttt{Nesterov} remains the best performer across all tested combinations of inner and outer optimizers (as seen in Figure~\ref{fig:all_optimizer}). This observation indicates that while momentum is beneficial in asynchronous \lsgd{} for language modeling, its effect is more pronounced in synchronous settings.

\begin{figure}[h!]
    \centering
    \includegraphics[width=0.5\textwidth]{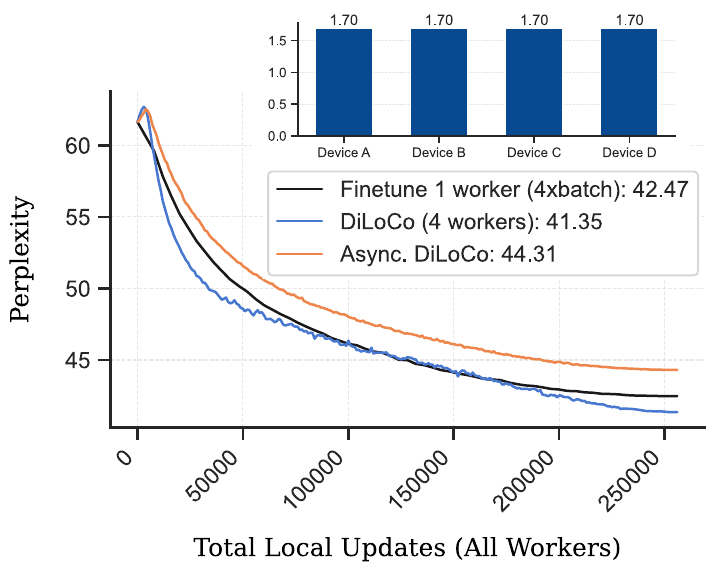}
    \caption{Async. DiLoCo with heterogeneous devices.}
    \label{fig:no_staleness_partial}
\end{figure}

\myparagraph{Is Staleness the Cause?} We further apply the asynchronous DiLoCo algorithm with homogeneous devices. By doing so, we maximally diminish the staled gradient problem in \lsgd{}, which refers to the fact that we are using an outdated outer gradient to update the server model. In particular, this means if we have $k$ workers, all of them will return the computed outer gradient back to the server at the same time. Therefore, the only staleness comes from the fact that we are sequentially applying the individual updates instead of aggregating them together and apply it once. Results are summarized in Figure~\ref{fig:no_staleness_partial}.

\myobs{Observation} Figure~\ref{fig:no_staleness_partial} reveals a notable finding: even with homogeneity among workers, asynchronous DiLoCo significantly lags behind its synchronous counterpart. This suggests that the \emph{inherent staleness} from sequentially applying simultaneous updates leads to considerable performance drops. To elucidate this effect, let's consider a scenario with $k=4$ workers providing identical outer gradients (denoted as $g$). The standard Nesterov momentum update is outlined in Equation~\eqref{eq:nesterov}. In a sequential application of pseudo-gradients:
\begin{equation}
\begin{split}
m_{t+1} &= \beta^4 m_t + (1 + \beta + \beta^2 + \beta^3) g \\
\theta_{t+1} &= \theta_t - \epsilon \big( (4 + 4\beta + 3\beta^2 + 2\beta^3 + \beta^4) g \\
&+ \beta^2 (1 + \beta + \beta^2 + \beta^3) m_t\big).
\end{split}
\end{equation}
From this, we observe that sequential application results in a more rapidly decaying momentum term but amplifies the actual change in parameter $\theta$. Consequently, a higher $\beta$ maintains more recent momentum but may lead to greater changes in parameters, and vice versa. Importantly, this imbalance cannot be simply rectified by reducing the learning rate.

\paragraph{Baselines} We consider several synchronous baselines from the literature, and their naive application to an asynchronous setting: \textbf{1)} Finetune 1 worker (4xbatch): This involves finetuning a single worker with a larger batch size, equating to synchronous SGD.
\textbf{2)} DiLoCo~\citep{douillard2023diloco}: This synchronous \lsgd{} method combines \texttt{AdamW} with \texttt{Nesterov}.
\textbf{3)} Async. DiLoCo: The asynchronous version of DiLoCo.

\begin{figure}[h!]
\centering
\includegraphics[width=0.5\textwidth]{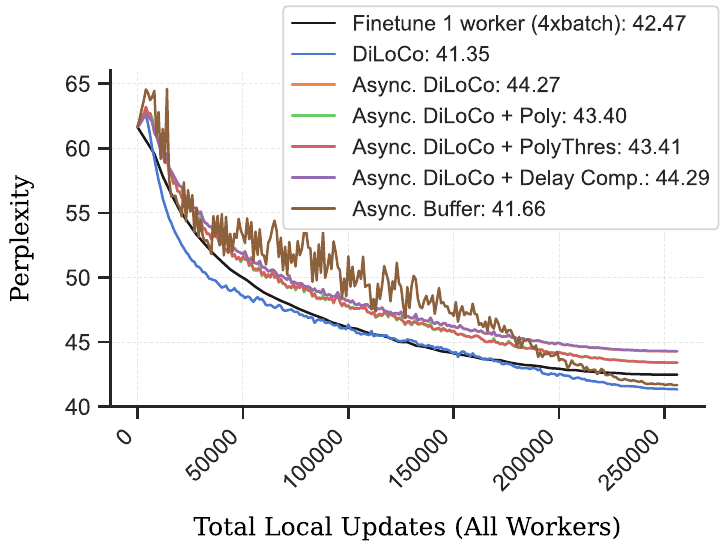}
\caption{Comparison of different asynchronous \lsgd{} approaches. Poly, PolyThres, and Delay Comp. barely improve the async. \lsgd{} performance. Async. Buffer significantly closes the gap between sync. and async. training, while introducing instability in early stage of training.}
\label{fig:async_baselines}
\end{figure}

\myparagraph{Existing Fixes} We investigated potential fixes from the asynchronous \lsgd{} literature to address observed challenges. The following methods were considered:
\textbf{1)} Async. DiLoCo + Poly~\citep{xie2019asynchronous}: Extends Async. DiLoCo by downweighting the pseudo-gradient with $g \leftarrow (1 + \text{staleness})^{-0.5} g$.
\textbf{2)} Async. DiLoCo + PolyThres: Adds a threshold to discard updates with staleness beyond 10.
\textbf{3)} Async. DiLoCo + Delay Comp.~\citep{zheng2017asynchronous}: Introduces delay compensation (Delay Comp.) to approximate true pseudo-gradients. Denote the gradient function at $\theta$ as $g(\theta)$, then the main idea of delay compensation is to approximate the true gradient $g(\theta_{t})$ by a stale gradient $g(\theta_{t'})$ ($t' < t$) with the first-order Taylor approximation, e.g., $g(\theta_t) \approx g(\theta_{t'}) + \nabla g(\theta_{t'})(\theta_t - \theta_{t'})$. In practice, the Hessian $\nabla g(\theta_{t'})$ is approximated by diagonalized gradient outer product, e.g., $\nabla g(\theta_{t'}) \approx \lambda g(\theta_{t'}) \odot g(\theta_{t'})$, where $\lambda \in \mathbb{R}^+$ is a hyperparameter. In our setting, we apply the delay compensation technique to pseudogradients instead of gradients.
\textbf{4)} Async. Buffer: Accumulates and averages all gradients in a First-In, First-Out fashion before applying Nesterov updates; a variation of the original FedBuff algorithm~\citep{nguyen2022federated}, using \texttt{AdamW}+\texttt{Nesterov}.
The results are provided in Figure~\ref{fig:async_baselines}.

\myobs{Observation} Polynomial discounting of the pseudo-gradient shows marginal benefits. Thresholding and delay compensation techniques don’t offer much improvements. Again, the fact that delay compensation is not working well points out the difference between asynchronous SGD and asynchronous \lsgd{}. The Async. Buffer method excels at convergence but exhibits instability early in training. Crucially, \emph{none} of the methods match the performance of the synchronous DiLoCo baseline.

\section{Proposed Solutions} \label{sec:method}
In addressing the optimization challenges outlined in Section~\ref{sec:challenge}, we developed two strategies.
\paragraph{Delayed Nesterov Update}
\label{sec:dn}
Notably, the Async. Buffer method demonstrated promising performance (as shown in Figure~\ref{fig:async_baselines}). Additionally, our analysis revealed that asynchronous training with \texttt{AdamW}+\texttt{SGD}, sans outer momentum, outperforms synchronous methods (Figure~\ref{fig:all_optimizer}). Based on these insights, we propose the \emph{Delayed Nesterov} (DN) strategy, which represents the \myfn{sync}() function in Algorithm~\ref{alg:async_pipeline}. This approach involves using the \texttt{Nesterov} update intermittently—every $N$ server updates. Between \texttt{Nesterov} updates, we aggregate pseudo-gradients in a buffer $\Delta$ and update the model parameters using gradient descent (or gradient descent plus a small fraction of the old momentum). To balance gradient and momentum-based descent, we introduce a parameter $c \in [0, 1/N]$. A $c$ value of 0 indicates pure gradient descent between \texttt{Nesterov} updates, while $c$ equal to 1 evenly distributes the momentum term over $N$ updates. The specifics of this algorithm are detailed in Algorithm~\ref{alg:delayed_nesterov}. Unlike the Async. Buffer~\citep{nguyen2022federated}, which updates model parameters only once in $N$ periods, the Delayed Nesterov continuously updates using gradients, incorporating a fraction of the old momentum, and updates the momentum term once every $N$ server updates.

\begin{algorithm}[t!]
\caption{Delayed Nesterov Update.} \label{alg:delayed_nesterov}
\begin{algorithmic}
\Require Initial model parameter $\theta_0$
\Require Momentum decay $\beta \in (0, 1)$
\Require Momentum activation $c \in [0, 1/N]$
\State \Comment{\textcolor{gray}{default to $c=0$}}
\Require Buffer size $N$
\State $t = 0$
\State $m_0 = 0$ \Comment{\textcolor{gray}{momentum}}
\State $\Delta = 0$ \Comment{\textcolor{gray}{aggregated gradient}}
\While{not finished}
    \State Receive the pseudo-gradient $g_t$ 
    \State \Comment{\textcolor{gray}{sync. step in Alg.~\ref{alg:async_pipeline}}}.
    \State $\Delta \leftarrow \Delta + g_t$
    \If{$(t+1) ~\%~ N == 0$}
        \State $m_{t + 1} \leftarrow \beta m_{t} + \Delta / N $
        \State $\theta_{t + 1} \leftarrow \theta_{t} - \epsilon \big((1 - cN + c) \beta m_{t + 1} + g_t / N\big)$
        \State $\Delta = 0$
    \Else
        \State $m_{t + 1} \leftarrow m_{t}$ \Comment{\textcolor{gray}{delay momentum update}}
        \State $\theta_{t + 1} \leftarrow \theta_{t} - \epsilon \big(c \beta m_{t + 1} + g_t / N\big)$
    \EndIf
    \State $t \leftarrow t + 1$
\EndWhile
\end{algorithmic}
\end{algorithm}

\paragraph{Dynamic Local Updates}
\label{sec:dylu}
The Delayed Nesterov approach addresses the momentum challenge in the \texttt{OuterOpt} by buffering pseudo-gradients and strategically delaying momentum updates. An alternative perspective considers synchronous training as a solution, where pseudo-gradients from all workers are synchronized. However, the diversity in device capabilities often hinders simultaneous pseudo-gradient returns, if each worker executes the same number of local training steps. A viable workaround involves customizing local training steps (e.g., $w$.steps) based on the processing speed of each device. In particular, denote $v(w)$ as the training speed (in terms of steps per second) for worker $w$, we can compute a worker's desired training steps as:
\begin{equation}
    w.\text{step} = \bigg\lfloor \frac{v(w)}{\max_{w' \in \mathcal{W}} v(w')} H \bigg\rfloor,
    \label{eq:DyLU}
\end{equation}
where $H$ denotes the number of local training steps the fastest worker runs and $\lfloor x \rfloor$ denotes the largest integer not greater than $x$.\footnote{Here, we implicitly assumes the device speeds are known a priori. If this is not the case, it is straightforward to estimate the device speed based on empirical observations.} We name this approach the Dynamic Local Updates (DyLU). This adjustment allows slower workers to execute fewer steps, aligning the completion times across different workers. Incorporating a grace period for model synchronization in this setup further reduces the impact of stale gradients, improving overall performance.

\section{A Minimal Toy Example} \label{sec:toy}
\begin{figure}[h!]
    \centering
    \includegraphics[width=\columnwidth]{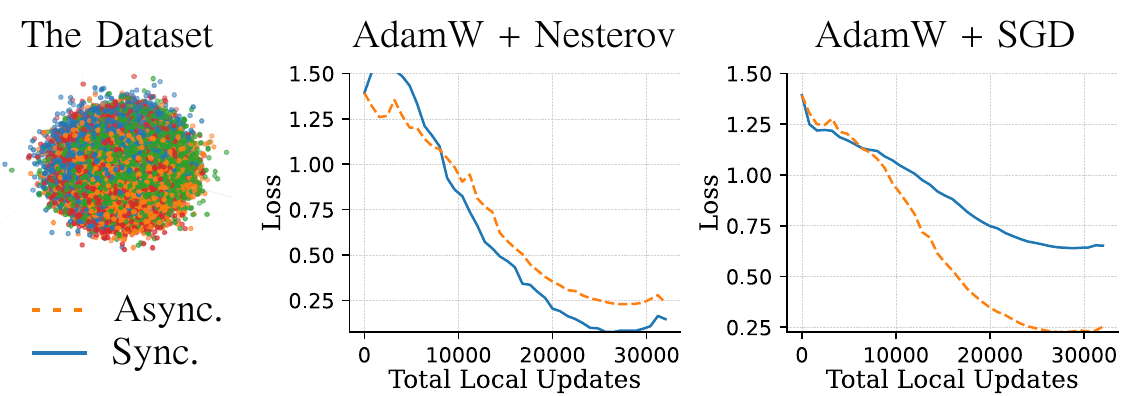}
    \caption{Replicating the optimization challenge on the toy example. \textbf{Left}: the dataset consists of a mixture of mixtures of Gaussians. \textbf{Right}: Async. \lsgd{} performs worse/better than sync. \lsgd{} when using \texttt{AdamW}+\texttt{Nesterov}/\texttt{AdamW}+\texttt{SGD}.}
    \label{fig:toy}
\end{figure}

For the convenience of future research and quick prototyping of new ideas, we present a minimal toy example that replicates the observed optimization challenge in asynchronous \lsgd{} (See Figure~\ref{fig:toy}).\footnote{Please check the Colab at \url{https://github.com/google-deepmind/asyncdiloco}} The task is to perform classification on a mixture of mixtures of Gaussian data.

\myobs{Observation} Comparing Figure~\ref{fig:toy} to Figure~\ref{fig:nesterov_or_not}, we observe that the toy example demonstrate the same optimization challenge.

\section{Experiments} \label{sec:experiments}
This section details experiments conducted to assess the efficacy of our two proposed methods, Delayed Nesterov (DN) and Dynamic Local Updates (DyLU). Additionally, ablation studies explore the effectiveness of these methods as we vary the number of workers and model sizes.

\myparagraph{Evaluating Delayed Nesterov (DN) and Dynamic Local Updates (DyLU)} Figure~\ref{fig:async_delayed_nesterov_very} compares asynchronous \lsgd{} with DN and DyLU against baselines such as single worker finetuning and DiLoCo, using the same setup as in Figure~\ref{fig:async_baselines}.

\myobs{Observation} The results demonstrate that DN combined with DyLU significantly reduces perplexity, surpassing the synchronous DiLoCo's performance over updates. Additionally, DN+DyLU outperforms in terms of time efficiency, avoiding delays from slower workers.

\begin{figure*}[t!]
    \centering
    \includegraphics[width=1\textwidth]{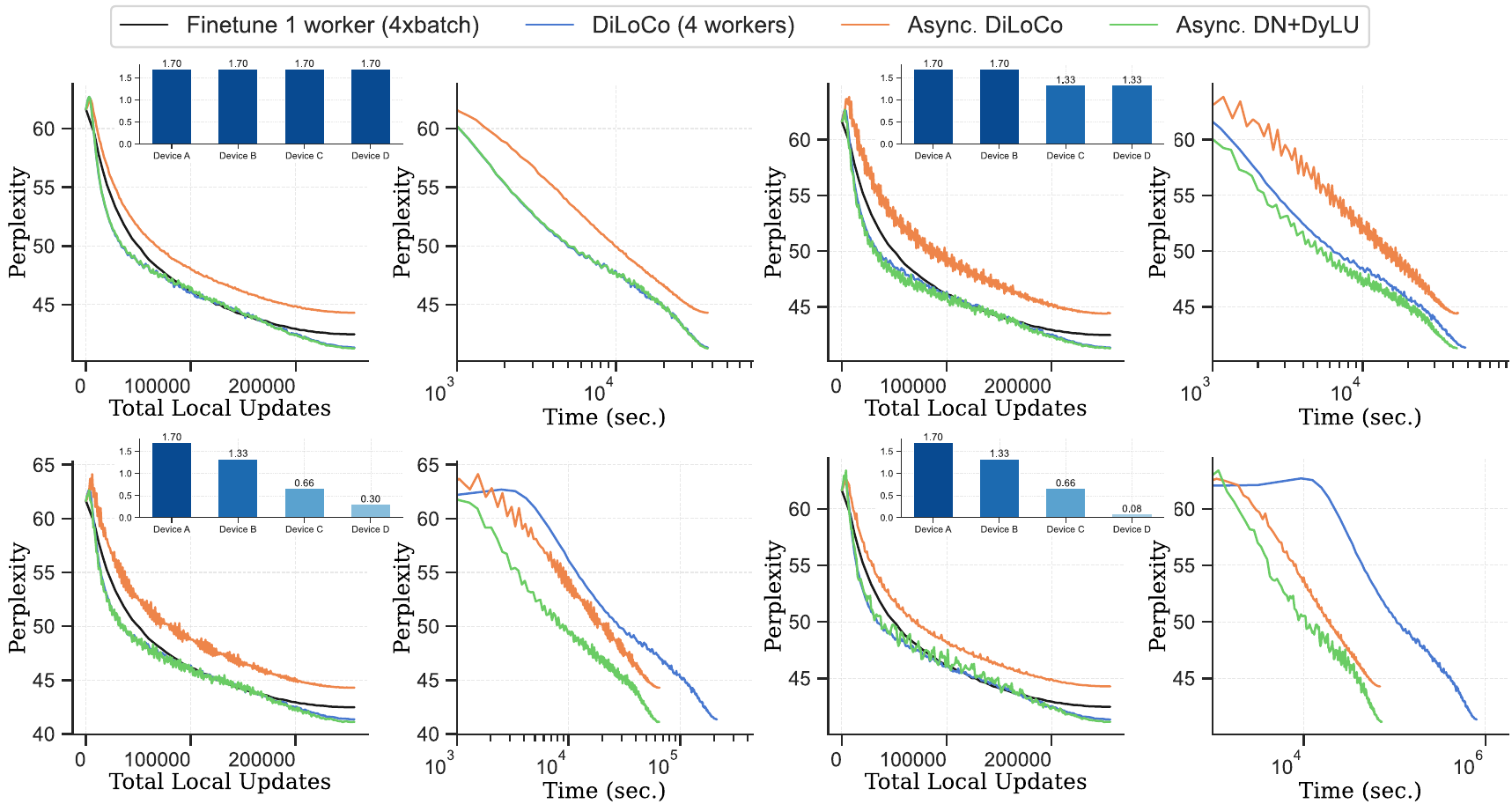}
    \caption{Varying the heterogeneity in devices.}
    \label{fig:async_vary_heterogeneity}
\end{figure*}

\myparagraph{Assessing Different Levels of Worker Heterogeneity} We examine how both the proposed DN+DyLU method and vanilla asynchronous DiLoCo fare under varying degrees of worker device heterogeneity, as shown in Figure~\ref{fig:async_vary_heterogeneity} (perplexity curve) and Table~\ref{tab:async_vary_heterogeneity} (final perplexity).

\begin{table}[h!]
\centering
\resizebox{1.0\columnwidth}{!}{%
%\vspace*{-0.3cm}
\begin{tabular}{@{}l c c c c}
\toprule
Level of heterogeneity & no & slight & moderate & very  \\
\midrule
Pretrained (24K)                         & 61.64 & 61.64 & 61.64 & 61.64 \\
Finetune ($4\times$ batch size)          & 42.47 & 42.47 & 42.47 & 42.47 \\
DiLoCo~\citep{douillard2023diloco}       & 41.35 & 41.35 & 41.35 & 41.35 \\
\midrule
Async. DiLoCo                            & 44.27 & 44.38 & 44.29 & 44.27 \\
Async. DN + DyLU (ours)                  & \textbf{41.27} & \textbf{41.27} & \textbf{41.09} & \textbf{41.13} \\
\bottomrule
\end{tabular}
}
\caption{Varying the level of worker heterogeneity (\textbf{top-left}, \textbf{top-right}, \textbf{bottom-left}, and \textbf{bottom-right} of Figure~\ref{fig:async_vary_heterogeneity} correspond to \textbf{no}, \textbf{slight}, \textbf{moderate}, and \textbf{very}, respectively).}
\label{tab:async_vary_heterogeneity}
\end{table}

\myobs{Observation} DN+DyLU consistently excels across all heterogeneity levels.\footnote{We notice that Async. DN+DyLU performs slightly better than DiLoCo when there is no heterogeneity, this is due to the numerical error, as the two methods reduce to the same and the training curves match almost perfectly.} Interestingly, even with homogeneous devices, vanilla asynchronous DiLoCo struggles, suggesting that the issue partly lies in the sequential application of pseudogradients. This underscores the importance of delayed momentum updates. Additionally, a periodic oscillation in performance is observed in certain device groupings, further highlighting the lack of robustness of the original asynchronous algorithm.

\begin{figure*}[t!]
    \centering
    \includegraphics[width=1\textwidth]{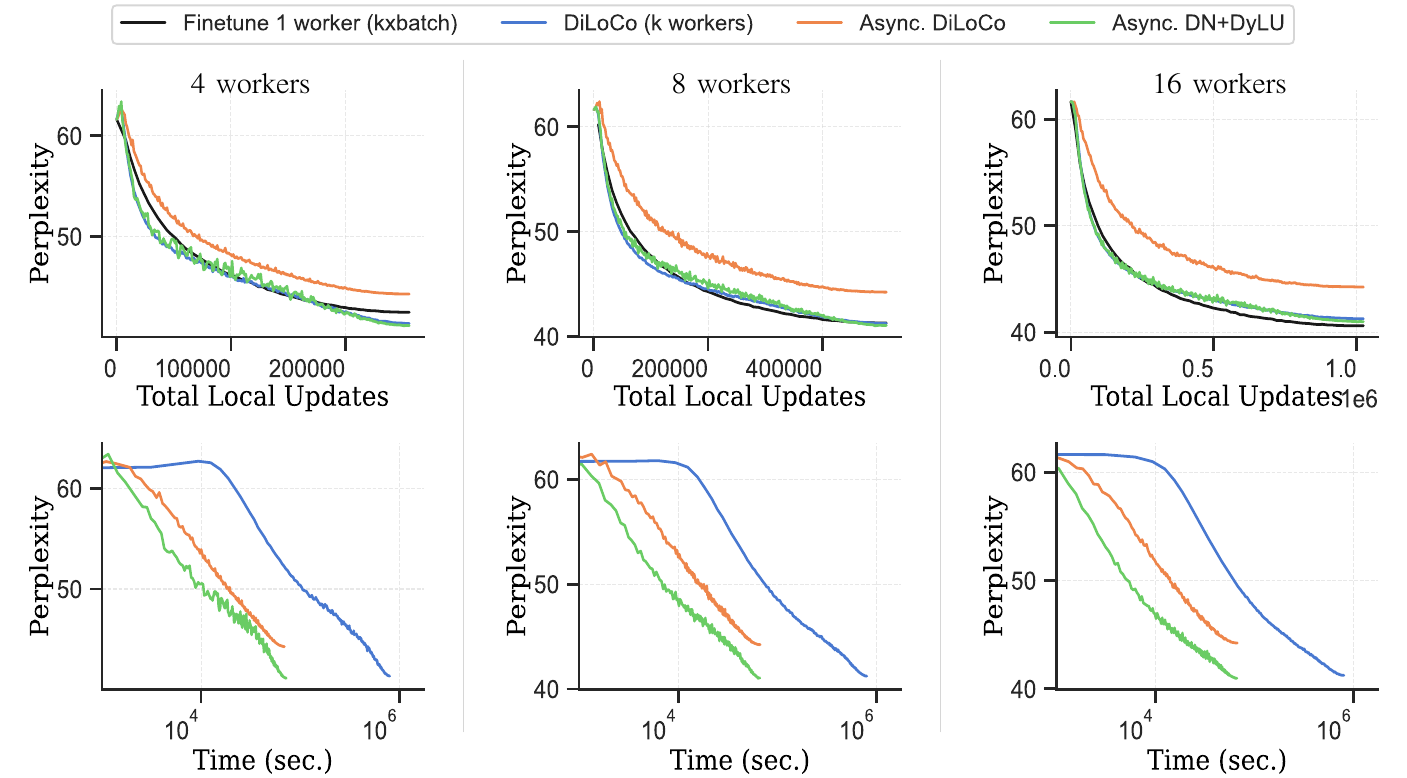}
    \caption{Varying the number of workers.}
    \label{fig:async_workers}
\end{figure*}

\myparagraph{Ablation with Different Numbers of Workers} We apply DN+DyLU while varying the number of workers (4, 8, 16) using a 20M model, with results summarized in Figure~\ref{fig:async_workers} (perplexity curve) and Table~\ref{tab:async_workers} (final perplexity).

\begin{table}[h!]
\centering
\resizebox{1.0\columnwidth}{!}{%
%\vspace*{-0.3cm}
\begin{tabular}{@{}l c c c}
\toprule
Number of workers $k$ & 4 & 8 & 16 \\
\midrule
Pretrained (24K)                         & 61.64 & 61.64 & 61.64 \\
Finetune ($k\times$ batch size)          & 42.47 & 41.28 & \textbf{40.60} \\
DiLoCo~\citep{douillard2023diloco}       & 41.35 & 41.23 & 41.25 \\
\midrule
Async. DiLoCo                            & 44.27 & 44.23 & 44.23 \\
Async. DN + DyLU (ours)                  & \textbf{41.13} & \textbf{41.02} & 40.98 \\
\bottomrule
\end{tabular}
}
\caption{Varying the number of workers.}
\label{tab:async_workers}
\end{table}

\myobs{Observation} As the number of workers increases, the benefit of \lsgd{} training diminishes. Notably, with 16 workers, single worker finetuning (16x batch size) shows the best performance over updates. Yet, DN+DyLU closely aligns with synchronous DiLoCo in performance, demonstrating its potential as a DiLoCo alternative in heterogeneous settings.

\begin{figure*}[t!]
    \centering
    \includegraphics[width=1\textwidth]{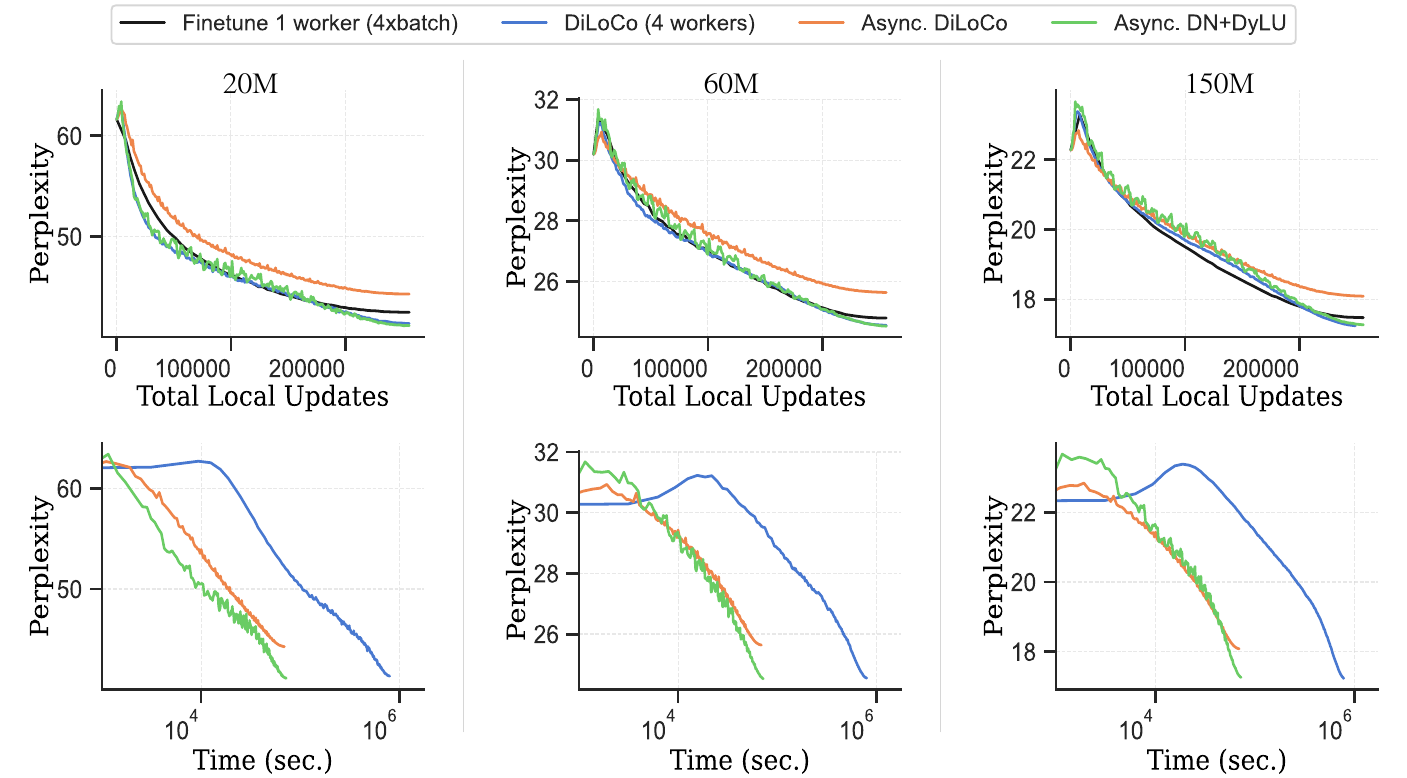}
    \caption{Varying the model size.}
    \label{fig:async_model_sizes}
\end{figure*}

\myparagraph{Ablation with Different Model Sizes} Lastly, we apply DN+DyLU to models of varying sizes (20M, 60M, 150M), with results summarized in Figure~\ref{fig:async_model_sizes} (perplexity curve) and Table~\ref{tab:async_model_sizes} (final perplexity).

\begin{table}[h!]
\centering
\resizebox{1.0\columnwidth}{!}{%
%\vspace*{-0.3cm}
\begin{tabular}{@{}l c c c}
\toprule
Model size & 20M & 60M & 150M \\
\midrule
Pretrained (24K)                         & 61.64 & 30.19 & 22.80 \\
Finetune (4x batch size)                 & 42.47 & 24.80 & 17.47 \\
DiLoCo~\citep{douillard2023diloco}       & 41.35 & 24.55 & \textbf{17.23} \\
\midrule
Async. DiLoCo                            & 44.27 & 25.64 & 18.08 \\
Async. DN + DyLU (ours)                  & \textbf{41.13} & \textbf{24.53} & 17.26 \\
\bottomrule
\end{tabular}
}
\caption{Varying the model sizes.}
\label{tab:async_model_sizes}
\end{table}

\myobs{Observation} Both synchronous and asynchronous \lsgd{} methods outperform the approach of finetuning a single worker with an increased batch size. Notably, this advantage becomes more pronounced during the later stages of convergence, aligning with findings from previous research that highlight \lsgd{}'s superior generalization capabilities~\citep{gu2023and}. Additionally, our proposed DN+DyLU method demonstrates consistent efficacy across various model sizes. It's important to note that the performance disparity between synchronous and asynchronous DiLoCo does not diminish even as the model size increases.

\myparagraph{Ablation with Different $c$}
We apply $c \in \{0, 0.1\}$ in Async. DN+DyLU with varying $k$ (4, 8, 16) and model sizes (20M, 60M, 150M), with the 4 ``very" heterogeneous workers. This is because when the level of heterogeneity is small, using different $c$ will have smaller difference (e.g., when there is no heterogeneity, any $c$ results in the same algorithm). Results are summarized in Table~\ref{tab:async_c}.

\begin{table}[h!]
\centering
\resizebox{1.0\columnwidth}{!}{%
%\vspace*{-0.3cm}
\begin{tabular}{@{}l c c c}
\toprule
Number of workers $k$ & 4 & 8 & 16 \\
\midrule
Async. DN + DyLU ($c=0$)        & \textbf{41.13} & 41.02 & \textbf{40.98} \\ 
Async. DN + DyLU ($c=0.1$)      & 41.16 & \textbf{40.93} & 41.04 \\ 
\toprule
Model size & 20M & 60M & 150M \\
\midrule
Async. DN + DyLU ($c=0$)        & \textbf{41.13} & \textbf{24.53} & \textbf{17.26} \\     
Async. DN + DyLU ($c=0.1$)      & 41.16 & 24.69 & 17.27 \\ 
\bottomrule
\end{tabular}
}
\caption{Varying the $c \in \{0, 0.1\}$ in Algorithm~\ref{alg:delayed_nesterov}.}
\label{tab:async_c}
\end{table}

\myobs{Observation} Empirically, we observe no significant difference between $c=0$ and $c=0.1$, indicating that adding slight momentum at intermediate steps does not help too much. As a result, we set $c=0$ as the default value in Algorithm~\ref{alg:delayed_nesterov}, which corresponds to performing SGD updates between two consecutive Nesterov updates. Note that setting the value of $c$ does  not introduce any overhead to the overall algorithm.

\section{Related Work} \label{sec:related_work}
This section provides a concise overview of the literature on federated learning and local-SGD style distributed optimization, particularly focusing on their applications in asynchronous settings.

\paragraph{Local-SGD and Distributed Optimization} Local-SGD is a specific distributed optimization technique designed to reduce communication frequency~\citep{stich2018local, zhang2016parallel, bijral2016data, mcdonald2010distributed, coppola2015iterative, zinkevich2010parallelized}. The core principle of Local-SGD is to let each worker execute several local training iterations prior to engaging in global synchronization. This technique was later applied to the federated learning setting, leading to the development of the FedAvg method~\citep{mcmahan2017communication}, which aims to reduce communication overhead. Unlike Local-SGD, federated learning also addresses user privacy issues and typically involves heterogeneous devices. To further minimize communication overhead, FedOpt integrates adaptive optimization methods like SGD momentum and Adam~\citep{reddi2020adaptive}. However, as client/worker heterogeneity increases, the convergence rate often deteriorates. Methods like SCAFFOLD~\citep{karimireddy2020scaffold} and MIME~\citep{karimireddy2021breaking} have been introduced to adapt these optimization methods for heterogeneous environments.

\paragraph{Asynchronous Training} Asynchronous training was developed to mitigate the ``straggler effect" observed in synchronous distributed optimization, where learning efficiency is bottlenecked by the slowest worker~\citep{koh2006parallel, recht2011hogwild, dean2012large, lian2015asynchronous, lian2018asynchronous, diskin2021distributed}. A significant challenge in asynchronous optimization is the staled gradient problem, which occurs when an outdated gradient is applied to a recently updated model. Asynchronous SGD with delay compensation~\citep{zheng2017asynchronous} addresses this issue by approximating the true gradient using the old gradient. Asynchronous methods have also been explored in federated learning contexts~\citep{xie2019asynchronous}. Despite the challenge, asynchronous training has demonstrated success for language modeling as well~\citep{diskin2021distributed}, by using heterogeneous devices across the world.

\paragraph{Local-SGD for Language Modeling} The concept of local-SGD (or FedAvg) has previously been applied in the realm of language modeling. Cross-device federated learning, for instance, has been utilized to pretrain and fine-tune language models~\citep{hilmkil2021scaling,ro2022scaling,ryabinin2021moshpit,diskin2021distributedcollab,presser2020stub,borzunov2023petals}. More recently, DiLoCo has extended the local-SGD methodology to encompass larger language models, specifically proposing the use of AdamW + Nesterov momentum as the \texttt{InnerOpt} + \texttt{OuterOpt} pairing. In asynchronous settings, the FedBuff~\citep{nguyen2022federated} algorithm buffers pseudogradients from clients, updating the server model only after accumulating a sufficient number of pseudogradients. TimelyFL~\citep{zhang2023timelyfl} aims to reduce asynchrony by allowing slower devices to train only parts of the model.
\section{Limitations} \label{sec:limitations}
This study, while comprehensive, has several limitations. First, we identify a significant optimization challenge linked to momentum updates in the \texttt{OuterOpt}, but the precise cause of this issue remains unclear. Understanding this challenge with robust theoretical backing presents an intriguing avenue for future research. Second, our empirical observations suggest that the advantages of the \lsgd{} method diminish with an increasing number of workers, a phenomenon whose underlying reasons are yet to be understood. This issue currently hinders the scalability of asynchronous \lsgd{}. Finally, although our proposed method DN+DyLU shows improved empirical performance, it lacks formal theoretical convergence guarantees, an aspect that merits further investigation.
\section{Conclusion} \label{sec:conclusions}
This study presents a thorough examination of asynchronous \lsgd{} in language modeling.  Our central finding is that while momentum in the outer optimization loop is crucial, it may be less effective in asynchronous scenarios compared to synchronous ones when implemented naively. To bridge this gap, we introduce a novel approach, focusing on sporadic momentum updates using buffered pseudogradients, combined with continuous stochastic pseudogradient updates. Furthermore, our research reveals that tailoring local training steps to each worker's computational speed is not only a straightforward but also an efficient strategy to enhance performance.

However, there is much work to be done.  In the standard (as opposed to the ``local'') gradient descent setting, the optimal batch size in terms of decreasing loss as quickly as possible in terms of number of weight updates is not usually ``as large as possible''.  In our view, similarly, there is hope for asynchronous \lsgd{} methods that give better results per local update than synchronous \lsgd{}.

\section*{Acknowledgements}
We would like to thank Adam Fisch for his valuable feedback.

% Bibliography components
\bibliographystyle{plainnat}
\nobibliography*
\bibliography{refs}

\begin{thebibliography}{32}
\providecommand{\natexlab}[1]{#1}
\providecommand{\url}[1]{\texttt{#1}}
\expandafter\ifx\csname urlstyle\endcsname\relax
  \providecommand{\doi}[1]{doi: #1}\else
  \providecommand{\doi}{doi: \begingroup \urlstyle{rm}\Url}\fi

\bibitem[Bijral et~al.(2016)Bijral, Sarwate, and Srebro]{bijral2016data}
Avleen~S Bijral, Anand~D Sarwate, and Nathan Srebro.
\newblock On data dependence in distributed stochastic optimization.
\newblock \emph{arXiv preprint arXiv:1603.04379}, 2016.

\bibitem[Borzunov et~al.(2022)Borzunov, Baranchuk, Dettmers, Ryabinin, Belkada, Chumachenko, Samygin, and Raffel]{borzunov2023petals}
Alexander Borzunov, Dmitry Baranchuk, Tim Dettmers, Max Ryabinin, Younes Belkada, Artem Chumachenko, Pavel Samygin, and Colin Raffel.
\newblock Petals: Collaborative inference and fine-tuning of large models.
\newblock \emph{arXiv preprint library}, 2022.

\bibitem[Coppola(2015)]{coppola2015iterative}
Gregory~Francis Coppola.
\newblock Iterative parameter mixing for distributed large-margin training of structured predictors for natural language processing.
\newblock 2015.

\bibitem[Dean et~al.(2012)Dean, Corrado, Monga, Chen, Devin, Mao, Ranzato, Senior, Tucker, Yang, et~al.]{dean2012large}
Jeffrey Dean, Greg Corrado, Rajat Monga, Kai Chen, Matthieu Devin, Mark Mao, Marc'aurelio Ranzato, Andrew Senior, Paul Tucker, Ke~Yang, et~al.
\newblock Large scale distributed deep networks.
\newblock \emph{Advances in neural information processing systems}, 25, 2012.

\bibitem[Diskin et~al.(2021{\natexlab{a}})Diskin, Bukhtiyarov, Ryabinin, Saulnier, Lhoest, Sinitsin, Popov, Pyrkin, Kashirin, Borzunov, Villanova~del Moral, Mazur, Kobelev, Jernite, Wolf, and Pekhimenko]{diskin2021distributedcollab}
Michael Diskin, Alexey Bukhtiyarov, Max Ryabinin, Lucile Saulnier, Quentin Lhoest, Anton Sinitsin, Dmitry Popov, Dmitry Pyrkin, Maxim Kashirin, Alexander Borzunov, Albert Villanova~del Moral, Denis Mazur, Ilia Kobelev, Yacine Jernite, Thomas Wolf, and Gennady Pekhimenko.
\newblock Distributed deep learning in open collaborations.
\newblock \emph{Advances in Neural Information Processing Systems (NeurIPS)}, 2021{\natexlab{a}}.

\bibitem[Diskin et~al.(2021{\natexlab{b}})Diskin, Bukhtiyarov, Ryabinin, Saulnier, Sinitsin, Popov, Pyrkin, Kashirin, Borzunov, Villanova~del Moral, et~al.]{diskin2021distributed}
Michael Diskin, Alexey Bukhtiyarov, Max Ryabinin, Lucile Saulnier, Anton Sinitsin, Dmitry Popov, Dmitry~V Pyrkin, Maxim Kashirin, Alexander Borzunov, Albert Villanova~del Moral, et~al.
\newblock Distributed deep learning in open collaborations.
\newblock \emph{Advances in Neural Information Processing Systems}, 34:\penalty0 7879--7897, 2021{\natexlab{b}}.

\bibitem[Douillard et~al.(2023)Douillard, Feng, Rusu, Chhaparia, Donchev, Kuncoro, Ranzato, Szlam, and Shen]{douillard2023diloco}
Arthur Douillard, Qixuan Feng, Andrei~A Rusu, Rachita Chhaparia, Yani Donchev, Adhiguna Kuncoro, Marc'Aurelio Ranzato, Arthur Szlam, and Jiajun Shen.
\newblock Diloco: Distributed low-communication training of language models.
\newblock \emph{arXiv preprint arXiv:2311.08105}, 2023.

\bibitem[Gu et~al.(2023)Gu, Lyu, Huang, and Arora]{gu2023and}
Xinran Gu, Kaifeng Lyu, Longbo Huang, and Sanjeev Arora.
\newblock Why (and when) does local sgd generalize better than sgd?
\newblock \emph{arXiv preprint arXiv:2303.01215}, 2023.

\bibitem[Hilmkil et~al.(2021)Hilmkil, Callh, Barbieri, S{\"u}tfeld, Zec, and Mogren]{hilmkil2021scaling}
Agrin Hilmkil, Sebastian Callh, Matteo Barbieri, Leon~Ren{\'e} S{\"u}tfeld, Edvin~Listo Zec, and Olof Mogren.
\newblock Scaling federated learning for fine-tuning of large language models.
\newblock In \emph{International Conference on Applications of Natural Language to Information Systems}, pages 15--23. Springer, 2021.

\bibitem[Hoffmann et~al.(2022)Hoffmann, Borgeaud, Mensch, Buchatskaya, Cai, Rutherford, de~Las~Casas, Hendricks, Welbl, Clark, Hennigan, Noland, Millican, van~den Driessche, Damoc, Guy, Osindero, Simonyan, Elsen, Rae, Vinyals, and Sifre]{hoffmann2022chinchilla}
Jordan Hoffmann, Sebastian Borgeaud, Arthur Mensch, Elena Buchatskaya, Trevor Cai, Eliza Rutherford, Diego de~Las~Casas, Lisa~Anne Hendricks, Johannes Welbl, Aidan Clark, Tom Hennigan, Eric Noland, Katie Millican, George van~den Driessche, Bogdan Damoc, Aurelia Guy, Simon Osindero, Karen Simonyan, Erich Elsen, Jack~W. Rae, Oriol Vinyals, and Laurent Sifre.
\newblock Training compute-optimal large language models.
\newblock \emph{Advances in Neural Information Processing Systems (NeurIPS)}, 2022.

\bibitem[Karimireddy et~al.(2020)Karimireddy, Kale, Mohri, Reddi, Stich, and Suresh]{karimireddy2020scaffold}
Sai~Praneeth Karimireddy, Satyen Kale, Mehryar Mohri, Sashank Reddi, Sebastian Stich, and Ananda~Theertha Suresh.
\newblock Scaffold: Stochastic controlled averaging for federated learning.
\newblock In \emph{International conference on machine learning}, pages 5132--5143. PMLR, 2020.

\bibitem[Karimireddy et~al.(2021)Karimireddy, Jaggi, Kale, Mohri, Reddi, Stich, and Suresh]{karimireddy2021breaking}
Sai~Praneeth Karimireddy, Martin Jaggi, Satyen Kale, Mehryar Mohri, Sashank Reddi, Sebastian~U Stich, and Ananda~Theertha Suresh.
\newblock Breaking the centralized barrier for cross-device federated learning.
\newblock \emph{Advances in Neural Information Processing Systems}, 34:\penalty0 28663--28676, 2021.

\bibitem[Koh et~al.(2006)Koh, George, Haftka, and Fregly]{koh2006parallel}
Byung-Il Koh, Alan~D George, Raphael~T Haftka, and Benjamin~J Fregly.
\newblock Parallel asynchronous particle swarm optimization.
\newblock \emph{International journal for numerical methods in engineering}, 67\penalty0 (4):\penalty0 578--595, 2006.

\bibitem[Lian et~al.(2015)Lian, Huang, Li, and Liu]{lian2015asynchronous}
Xiangru Lian, Yijun Huang, Yuncheng Li, and Ji~Liu.
\newblock Asynchronous parallel stochastic gradient for nonconvex optimization.
\newblock \emph{Advances in neural information processing systems}, 28, 2015.

\bibitem[Lian et~al.(2018)Lian, Zhang, Zhang, and Liu]{lian2018asynchronous}
Xiangru Lian, Wei Zhang, Ce~Zhang, and Ji~Liu.
\newblock Asynchronous decentralized parallel stochastic gradient descent.
\newblock In \emph{International Conference on Machine Learning}, pages 3043--3052. PMLR, 2018.

\bibitem[Lin et~al.(2018)Lin, Stich, Patel, and Jaggi]{lin2018don}
Tao Lin, Sebastian~U Stich, Kumar~Kshitij Patel, and Martin Jaggi.
\newblock Don't use large mini-batches, use local sgd.
\newblock \emph{arXiv preprint arXiv:1808.07217}, 2018.

\bibitem[Lin et~al.(2020)Lin, Stich, Patel, and Jaggi]{Lin2020_localsgd}
Tao Lin, Sebastian~U. Stich, Kumar~Kshitij Patel, and Martin Jaggi.
\newblock Don't use large mini-batches, use local sgd.
\newblock \emph{Proceedings of the International Conference on Learning Representations (ICLR)}, 2020.

\bibitem[McDonald et~al.(2010)McDonald, Hall, and Mann]{mcdonald2010distributed}
Ryan McDonald, Keith Hall, and Gideon Mann.
\newblock Distributed training strategies for the structured perceptron.
\newblock In \emph{Human language technologies: The 2010 annual conference of the North American chapter of the association for computational linguistics}, pages 456--464, 2010.

\bibitem[McMahan et~al.(2017)McMahan, Moore, Ramage, Hampson, and y~Arcas]{mcmahan2017communication}
Brendan McMahan, Eider Moore, Daniel Ramage, Seth Hampson, and Blaise~Aguera y~Arcas.
\newblock Communication-efficient learning of deep networks from decentralized data.
\newblock In \emph{Artificial intelligence and statistics}, pages 1273--1282. PMLR, 2017.

\bibitem[Nguyen et~al.(2022)Nguyen, Malik, Zhan, Yousefpour, Rabbat, Malek, and Huba]{nguyen2022federated}
John Nguyen, Kshitiz Malik, Hongyuan Zhan, Ashkan Yousefpour, Mike Rabbat, Mani Malek, and Dzmitry Huba.
\newblock Federated learning with buffered asynchronous aggregation.
\newblock In \emph{International Conference on Artificial Intelligence and Statistics}, pages 3581--3607. PMLR, 2022.

\bibitem[Presser(2020)]{presser2020stub}
Shawn Presser.
\newblock Swarm training, 2020.
\newblock URL \url{https://battle.shawwn.com/swarm-training-v01a.pdf}.

\bibitem[Raffel et~al.(2020)Raffel, Shazeer, Roberts, Lee, Narang, Matena, Zhou, Li, and Liu]{c4}
Colin Raffel, Noam Shazeer, Adam Roberts, Katherine Lee, Sharan Narang, Michael Matena, Yanqi Zhou, Wei Li, and Peter~J. Liu.
\newblock Exploring the limits of transfer learning with a unified text-to-text transformer.
\newblock \emph{Journal of Machine Learning Research}, 2020.

\bibitem[Recht et~al.(2011)Recht, Re, Wright, and Niu]{recht2011hogwild}
Benjamin Recht, Christopher Re, Stephen Wright, and Feng Niu.
\newblock Hogwild!: A lock-free approach to parallelizing stochastic gradient descent.
\newblock \emph{Advances in neural information processing systems}, 24, 2011.

\bibitem[Reddi et~al.(2020)Reddi, Charles, Zaheer, Garrett, Rush, Kone{\v{c}}n{\`y}, Kumar, and McMahan]{reddi2020adaptive}
Sashank Reddi, Zachary Charles, Manzil Zaheer, Zachary Garrett, Keith Rush, Jakub Kone{\v{c}}n{\`y}, Sanjiv Kumar, and H~Brendan McMahan.
\newblock Adaptive federated optimization.
\newblock \emph{arXiv preprint arXiv:2003.00295}, 2020.

\bibitem[Ro et~al.(2022)Ro, Breiner, McConnaughey, Chen, Suresh, Kumar, and Mathews]{ro2022scaling}
Jae~Hun Ro, Theresa Breiner, Lara McConnaughey, Mingqing Chen, Ananda~Theertha Suresh, Shankar Kumar, and Rajiv Mathews.
\newblock Scaling language model size in cross-device federated learning.
\newblock \emph{arXiv preprint arXiv:2204.09715}, 2022.

\bibitem[Ryabinin et~al.(2021)Ryabinin, Gorbunov, Plokhotnyuk, and Pekhimenko]{ryabinin2021moshpit}
Max Ryabinin, Eduard Gorbunov, Vsevolod Plokhotnyuk, and Gennady Pekhimenko.
\newblock Moshpit sgd: Communication-efficient decentralized training on heterogeneous unreliable devices.
\newblock \emph{Advances in Neural Information Processing Systems}, 34:\penalty0 18195--18211, 2021.

\bibitem[Stich(2018)]{stich2018local}
Sebastian~U Stich.
\newblock Local sgd converges fast and communicates little.
\newblock \emph{arXiv preprint arXiv:1805.09767}, 2018.

\bibitem[Xie et~al.(2019)Xie, Koyejo, and Gupta]{xie2019asynchronous}
Cong Xie, Sanmi Koyejo, and Indranil Gupta.
\newblock Asynchronous federated optimization.
\newblock \emph{arXiv preprint arXiv:1903.03934}, 2019.

\bibitem[Zhang et~al.(2016)Zhang, De~Sa, Mitliagkas, and R{\'e}]{zhang2016parallel}
Jian Zhang, Christopher De~Sa, Ioannis Mitliagkas, and Christopher R{\'e}.
\newblock Parallel sgd: When does averaging help?
\newblock \emph{arXiv preprint arXiv:1606.07365}, 2016.

\bibitem[Zhang et~al.(2023)Zhang, Gao, Lee, Zhang, and Avestimehr]{zhang2023timelyfl}
Tuo Zhang, Lei Gao, Sunwoo Lee, Mi~Zhang, and Salman Avestimehr.
\newblock Timelyfl: Heterogeneity-aware asynchronous federated learning with adaptive partial training.
\newblock In \emph{Proceedings of the IEEE/CVF Conference on Computer Vision and Pattern Recognition}, pages 5063--5072, 2023.

\bibitem[Zheng et~al.(2017)Zheng, Meng, Wang, Chen, Yu, Ma, and Liu]{zheng2017asynchronous}
Shuxin Zheng, Qi~Meng, Taifeng Wang, Wei Chen, Nenghai Yu, Zhi-Ming Ma, and Tie-Yan Liu.
\newblock Asynchronous stochastic gradient descent with delay compensation.
\newblock In \emph{International Conference on Machine Learning}, pages 4120--4129. PMLR, 2017.

\bibitem[Zinkevich et~al.(2010)Zinkevich, Weimer, Li, and Smola]{zinkevich2010parallelized}
Martin Zinkevich, Markus Weimer, Lihong Li, and Alex Smola.
\newblock Parallelized stochastic gradient descent.
\newblock \emph{Advances in neural information processing systems}, 23, 2010.

\end{thebibliography}

\section*{Supplementary Materials} \label{sec:supp}
\subsection{Implementation Details}
\label{sec:implem_details}

\begin{table}[h!]
\centering
\resizebox{1.0\linewidth}{!}{%
%\vspace*{-0.3cm}
\begin{tabular}{@{}l|ccc@{}}
\toprule
Hyperparameter &  Value \\
\midrule
Inner learning rate & $0.1$ \\
Final inner learning rate & $0.0$, $\mathbf{0.000001}$, $0.0002$ \\
Number of warmup steps & $0$, $\mathbf{1{,}000}$\\
Weight decay & $0.1$\\
Batch Size & 128, 512 \\
Sequence length & $256$\\
\midrule
Outer Optimizer & SGD, SGDM, Nesterov, Adam, \textbf{delayed momentum SGD}\\
Inner Optimizer & SGD, \textbf{AdamW} \\
Outer learning rate & $0.03$, $0.3$, $\mathbf{0.1}$, $\mathbf{0.7}$\\
Async soup weight \citep{xie2019asynchronous} & $0.125$, $0.25$, $0.5$, $\mathbf{1.0}$ \\
Async soup method \citep{xie2019asynchronous} & \textbf{constant}, polynomial, svrg \\
Delay period & \textbf{4}, 8, 16 \\
Communication frequency $H$ & \textbf{50}, 100, 150 \\
Number of pretraining steps & $24,000$ \\
\bottomrule
\end{tabular}
}
\caption{\textbf{Optimization Hyperparameters} evaluated during in this work. Chosen values for main experiments are highlighted in bold.}
\label{tab:hyperparameters}
\end{table}

\begin{table}[t]
\centering
%\resizebox{1.0\linewidth}{!}{%
%\vspace*{-0.3cm}
\caption{\textbf{Model Configuration} for the three evaluated sizes. All are based on the transformer architecture, chinchilla-style \citep{hoffmann2022chinchilla}.}\vspace{3mm}
\begin{tabular}{@{}l|ccc@{}}
\toprule
Hyperparameter &  20M & 60M & 150M \\
\midrule
Number of layers & 6 & 3 & 12\\
Hidden dim & 256 & 896 & 896\\
Number of heads & 4 & 16 & 16\\
K/V size & 64 & 64 & 64\\
Vocab size & \multicolumn{3}{c}{$32{,}000$}\\
\bottomrule
\end{tabular}
%}
\label{tab:model_config}
\end{table}

\paragraph{Network Architecture}
We displayed in \autoref{tab:model_config} the architectural difference between the 20M, 60M, and 150M models. They are all transformer decoder-only, based on the Chinchilla family \citep{hoffmann2022chinchilla}.

\paragraph{Training Dataset} We consider a language modeling task on the C4 dataset, a dataset derived from Common Crawl~\citep{c4}. The total number of steps is set to $88{,}000$ for all models, with $24{,}000$ steps of pre-training done without any federated learning methods, akin to \textit{post Local-SGD} \citep{Lin2020_localsgd}.

\paragraph{Hyperparameters} In \autoref{tab:hyperparameters}, we outline the optimization hyperparameters considered for this study. %We detailled extensively the impact of each hyperparameters in \ref{sec:ablations}.

\paragraph{Inner Optimizer States} Following ~\citet{douillard2023diloco}, in all experiments, when worker B picks up the data shard worker A just finishes training on, we keep localy the \texttt{AdamW}'s optimizer state and don't communicate it between workers. Moreover, the same state is used from one round to another, without reset. The inner learning rate is scheduled through the entire training, across multiple rounds.

\subsection{Aync. Training  Pseudocode} \label{sec:pseudo}
In this section, we provide the pseudocode for the \myfn{train}() and \myfn{get\_worker}() functions in Algorithm~\ref{alg:async_pipeline}.

\begin{algorithm}[h!]
\begin{algorithmic}[1]
\Require Available workers $\mathcal{W}$
\Require Current server model $\theta$
\For{$w \in \mathcal{W}$}
\State Sample shard $\mathcal{D}'$ for $w$ (Eq.~\ref{eq:data_sample}).
\State $w$.local\_updates = \myfn{DyLU}($\mathcal{D}'$) (Eq.~\ref{eq:DyLU}).
\State Decide lr schedule ($w$.lr) (Eq.~\ref{eq:lr}).
\State $w$.update = \myfn{train\_worker}($w$, $\mathcal{D}'$, $\theta$).
\EndFor
\end{algorithmic}
\caption{\myfn{train}() in Algorithm~\ref{alg:async_pipeline}.}
\end{algorithm}

\begin{algorithm}[h!]
\begin{algorithmic}[1]
\Require Workers $\mathcal{W}$
\Require Grace period $\tau_\text{grace}$
\Require Start of the grace period $\tau_\text{sync}$.
    \If{all workers in $\mathcal{W}$ are not done}
        \State \textbf{return} null
    \Else
        \State $w$ = earliest completed worker in $\mathcal{W}$.
        \If{$w.\text{completed\_time} - \tau_\text{sync} \leq \tau_\text{grace}$}
            \State \textbf{return} $w$
        \Else 
            \State \textbf{return} null
        \EndIf
    \EndIf
\end{algorithmic}
\caption{\myfn{get\_worker}() in Algorithm~\ref{alg:async_pipeline}.}
\end{algorithm}

\end{document}